\documentclass[letterpaper, 10 pt, conference]{ieeeconf}  

\IEEEoverridecommandlockouts 

\usepackage{graphicx}
\usepackage{glossaries}
\usepackage{amsfonts}
\usepackage{amsmath}
\usepackage{amssymb}
\usepackage{caption}
\usepackage{tabularx}
\usepackage{algorithm2e}
\usepackage{textcomp}
\usepackage{tikz}   
\usepackage{graphicx}
\usepackage{subfigure}
\usepackage{subcaption}
\usepackage{algpseudocode}
\RestyleAlgo{ruled}

\usepackage{amsmath}

\DeclareMathOperator*{\argmin}{arg\,min}
\newcommand{\mb}[1]{\mathbf{#1}}

\newcommand{\de}[1]{\text{d}#1}

\newcommand {\bs}[1]{\boldsymbol{#1}}

\newcommand{\T}{^{\mathrm{T}}}  

\newacronym{mra}{MRA}{Magnetic Resonance Angiography}
\newacronym{mri}{MRI}{Magnetic Resonance Imaging}
\newacronym{cta}{CTA}{Computed Tomography Angiography}
\newacronym{ct}{CT}{Computed Tomography Angiography}
\newacronym{rrt}{RRT}{Rapidly exploring Random Tree}
\newacronym{lcca}{LCCA}{Left Common Carotid Artery}
\newacronym{lvo}{LVO}{Large Vessel Occlusion}
\newacronym{lsa}{LSA}{Left Subclavian Artery}

\newcommand*\circled[1]{\raisebox{.5pt}{\textcircled{\raisebox{-.9pt} {#1}}}}

\begin{document}

\title{\LARGE \bf Towards Autonomous Navigation of Neuroendovascular Tools for Timely Stroke Treatment via Contact-aware Path Planning}

\author{Aabha Tamhankar, Giovanni~Pittiglio
\thanks{FuTURE Lab, Department of Robotics Engineering, Worcester Polytechnic Insitute (WPI), Worcester, MA 01605, USA. Email: {\tt\small \{astamhankar, gpittiglio\}@wpi.edu}
\newline This work was supported by the Worcester Polytechnic Institute (WPI), Department of Robotics Engineering.}%
}

\maketitle

\begin{abstract}
In this paper, we propose a model-based contact-aware motion planner for autonomous navigation of neuroendovascular tools in acute ischemic stroke. The planner is designed to find the optimal control strategy for telescopic pre-bent catheterization tools such as guidewire and catheters, currently used for neuroendovascular procedures. A kinematic model for the telescoping tools and their interaction with the surrounding anatomy is derived to predict tools steering. By leveraging geometrical knowledge of the anatomy, obtained from pre-operative segmented 3D images, and the mechanics of the telescoping tools, the planner finds paths to the target enabled by interacting with the surroundings. We propose an actuation platform for insertion and rotation of the telescopic tools and present experimental results for the navigation from the base of the descending aorta to the \gls{lcca}. We demonstrate that, by leveraging the pre-operative plan, we can consistently navigate the \gls{lcca} with 100\% success of over 50 independent trials. We also study the robustness of the planner towards motion of the aorta and errors in the initial positioning of the robotic tools. The proposed plan can successfully reach the \gls{lcca} for rotations of the aorta of up to 10$^\circ$, and displacement of up to 10\,mm, on the coronal plane. 
\glsresetall
\end{abstract}


\IEEEpeerreviewmaketitle
\section{Introduction}
\label{sec:introduction}
Stroke is the second leading cause of death worldwide, accounting for 11.6\% of all deaths in 2019 \cite{Rai2023UpdatedCare}. Every 3 minutes and 14 seconds, someone dies of stroke \cite{2023StrokeFacts.} and 50\% of survivors become chronically disabled \cite{Donkor2018StrokeLife.} with physical, cognitive, speech, and other impairments. 87\% of all strokes occur when a blood clot obstructs the blood flow – or ischemia. \glspl{lvo} are ischemic events due to blockage of large vessels of the brain and affect 295,000 Americans yearly \cite{2023StrokeFacts.}. While Intravenous (IV) tissue Plasminogen Activator (tPA) – clot-busting medication - can generally restore blood flow in smaller vessels, mechanical thrombectomy is required for \glspl{lvo}. Mechanical thrombectomy after IV tPA restores blood flow in up to 80\% of \gls{lvo} cases \cite{LeeH.Schwamm2017EndovascularOutcomes}, while sole IV tPA recanalization results may be as low as 12\% \cite{LeeH.Schwamm2017EndovascularOutcomes}. While IV tPA administration does not require specialized training, thrombectomy is a minimally invasive procedure performed by trained and experienced neurointerventional units. 

\begin{figure}
    \centering
    \includegraphics[width=\columnwidth]{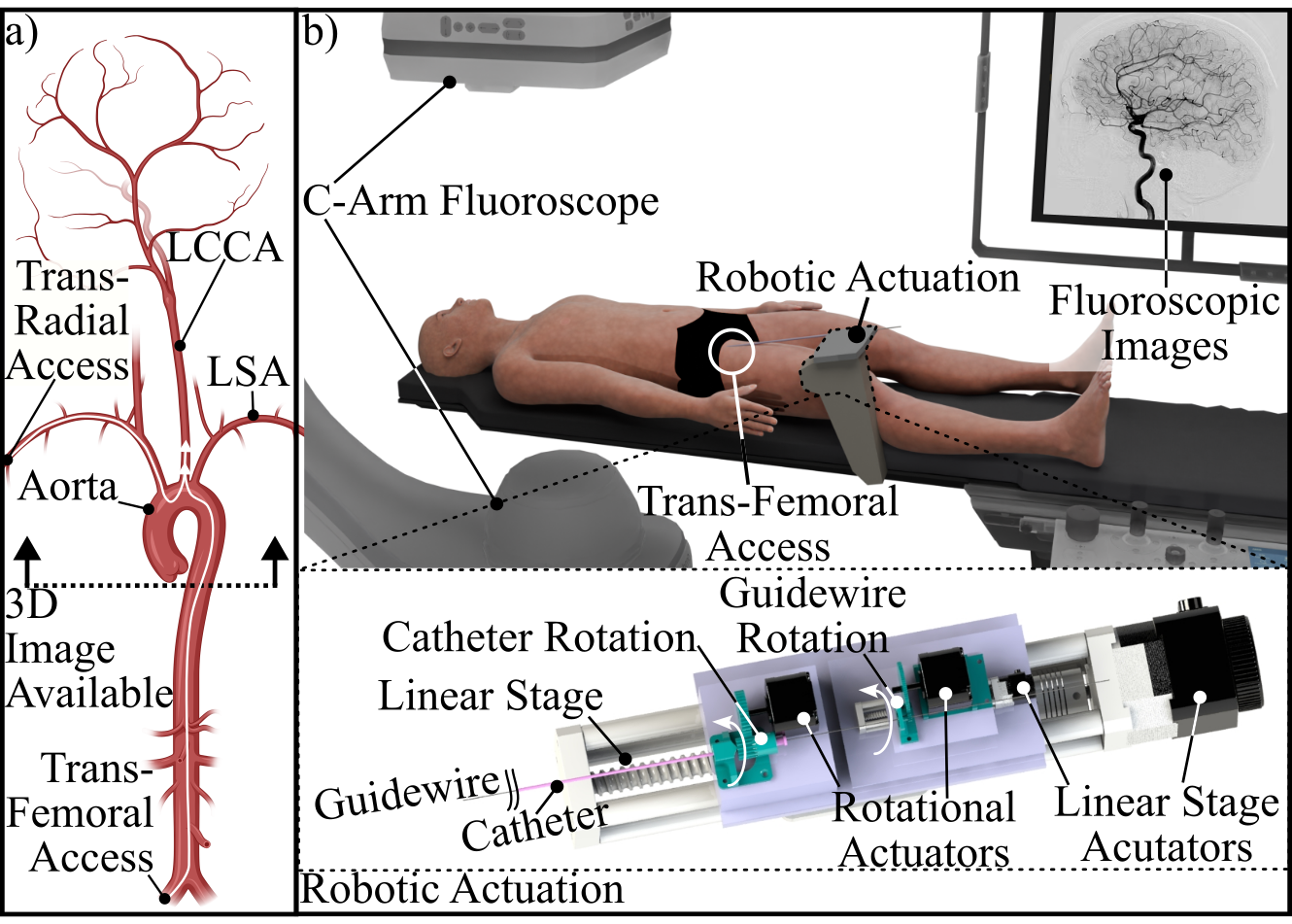}
    \caption{Description of robotic platform for autonomous navigation of neuroendovascular catheterization tools.}
    \label{fig:platform}
\end{figure}

A study from 2004 indicates that, in the USA, 82\% of patients live more than 3 hours away from centers capable of performing \gls{lvo} treatments \cite{Peterman2022GeospatialStates} and more recent studies show that rural-urban discrepancies are aggravating \cite{Hammond2020Urban-RuralMortality}. Time to treatment is critical: each hour of delay before mechanical thrombectomy results in a greater degree of disability \cite{Powers2019GuidelinesAssociation}. Time delays are also considered amongst the reasons for which we observe 20\% higher stroke mortality in patients from rural areas, compared to urban centers \cite{Llanos-Leyton2022DisparitiesStroke}. 

When a patient with stroke symptoms is admitted to an emergency unit, the nature of the stroke is investigated using \gls{ct} and/or \gls{mri} scans. If signs of ischemia are found, the patient is transferred to specialized medical centers. They perform a 3D scan, generally \gls{mra} or \gls{cta}, to visualize the arterial network from the aortic arch up to the brain vessels \cite{vanderZijden2019CurrentClinician}, and plan the navigation of telescoping flexible tools from the femoral (leg) or radial artery (arm) to the target location (Fig. \ref{fig:platform}). Once the site of flow blockage is reached, stent retrieval and/or aspiration are used to remove the blood clot.

Highly skilled interventional radiologists select combinations of telescoping guidewires and catheters (Fig. \ref{fig:model}) and navigate to the clot using information obtained from pre-operative \gls{cta} or \gls{mra}, and real-time fluoroscopy. They train in predicting how tools-anatomy interaction affects steering in the arteries, as they read images and insert and rotate guidewire and catheter. They integrate knowledge of landmarks in the anatomy, information gathered through \gls{cta} or \gls{mra} and partial information from fluoroscopy. This requires several cognitive skills, which are developed via extensive learning and require continuous practice so that they are not lost. However, only select large medical centers provide required training of their residents and have a workload which allows constant practice of their senior clinicians \cite{Hammond2020Urban-RuralMortality}.

Our goal is to develop a semi-autonomous robotic solution (Fig. \ref{fig:platform}) which, using pre-operative \gls{cta} or \gls{mra} images, can navigate pre-bent telescopic tools to the target location. We envision its use in small and rural centers to reduce time to treatment and reperfusion. Clinical personnel will gain arterial access, introduce the robot until the aorta (Fig. \ref{fig:platform}a), and supervise its navigation to the target site with tele-guidance of remote expert neuroendovascular surgeons. 

Under the hypothesis that higher tip steerability would improve navigation capabilities, several actuation technologies such as cable- \cite{Lis2022DesignRobot, Abah2024Self-SteeringInterventions} and magnetic-based \cite{Kim2022TeleroboticManipulation, Dreyfus2024DexterousAccess, Brockdorff2024HybridApplications, Pittiglio2022Patient-SpecificEndoscopy, Pittiglio2023PersonalizedLungs, Dreyfus2024DexterousAccessb} have been investigated. While these enable higher dexterity and improved targeting, they are associated with higher development and production costs. In contrast, we propose minimal hardware development, and developed a cost-effective actuation platform (Fig. \ref{fig:platform}b) to actuate standard endovascular catheterization tools (telescoping guidewire and catheters). 

In this paper, we present a \emph{contact-aware path planning} strategy able to interpret 3D images of the arteries, eventually provided by pre-operative \gls{cta} or \gls{mra}, and command the robots actions to steer and advance in the anatomy by leveraging tools-anatomy interaction. This strategy is inspired by standard manual catheterization approaches which uses no tools steerability and require interacting with the vessels' walls to navigate. Inspired by \cite{Pittiglio2023ClosedRobots}, we introduce a novel model for flexible telescoping tools able to predict their shape when introduced in a anatomy of known geometry, gathered from \gls{cta} or \gls{mra}. This model was used to plan robotic navigation inside the vessels, using a \gls{rrt} algorithm.

In this work, we focus on the first step of neuroendovascular catheterization from trans-femoral access (Fig. \ref{fig:platform}a): navigating from the aortic arch to the carotid arteries. Notice that the path from femoral artery to aorta is relatively straight and does not require specific planning. We present experimental validation in a realistic phantom of the anatomy to show the ability of our autonomous robotic platform to navigate from the base of the aorta to the \gls{lcca}, while avoiding the \gls{lsa}. We show repeatability over 50 trials and robustness towards unexpected motions of the anatomy.

\section{Catheterization Tools Static Model}
\label{sec:model}
\begin{figure}
    \centering
    \includegraphics[width=1\columnwidth]{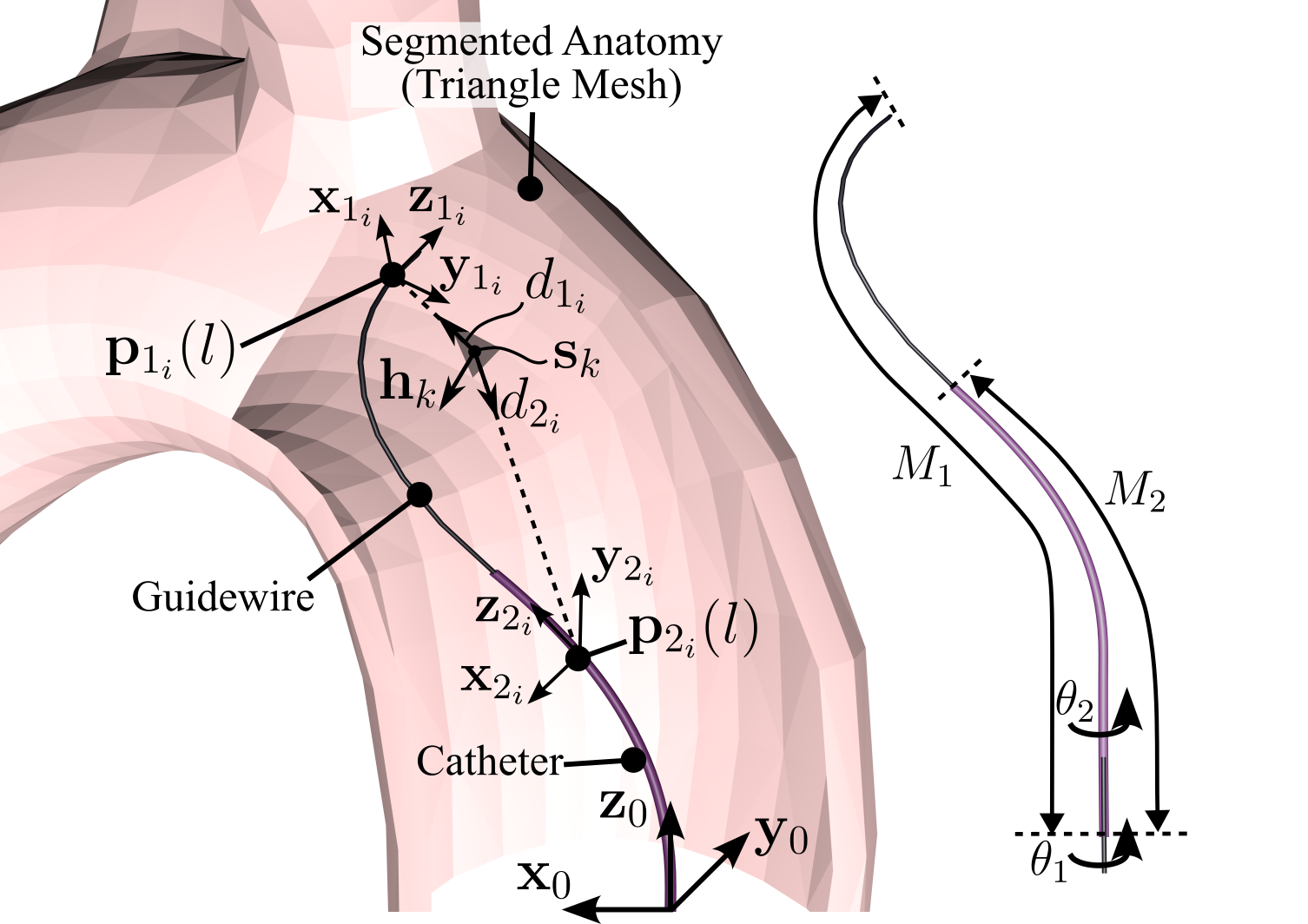}
    \caption{Schematic description of model parameters.}
    \label{fig:model}
\end{figure}

In Fig. \ref{fig:model}, we describe the telescoping tools for neuroendovascular catheterization. Generally, navigation is performed using combinations of a pre-bent guidewire and a pre-bent catheter whose insertion and rotation can be controlled. In the following, we describe how we model these tools and the interaction with the anatomy. The overall model is used to plan the autonomous navigation within the anatomy, as described in Section \ref{sec:planning}.

We model each tool's static equilibrium independently, as described in Section \ref{sec:model:single}, and treat the interaction between tools (Section \ref{sec:model:tools}) and with the anatomy (Section \ref{sec:model:anatomy}) as constraints to the static equilibrium of each tool. The solution to the constrained statics is described in Section \ref{sec:model:solution}.

In this work, we assume that frictional effects between tools and between tools and anatomy can be neglected. We also assume that the mechanics of the tools is dominated by their elastic behavior and that the anatomy is rigid when compared to the tools.

\subsection{Single Tool Model}
\label{sec:model:single}
Following \cite{Pittiglio2023ClosedRobots}, we describe each tool using infinitesimal segments each assumed to shape as an arc of a circle. The overall centerline of each tool, under this condition, follows the piece-wise constant curvature assumption. We describe the centerline of each segment as a function of their length $s_{j_i} \in [0, l]$, where $l \in \mathbb{R}$ is the length of the $i$-th segment, here assumed the same among segments and tools; the index $j$ indicates the tools number, $j = 1$ is the innermost. In this work, we consider two tools, as shown in Fig. \ref{fig:model}.

We associate a reference frame to each point along the tools' centerline $F_{j_i}(s_{j_i})=\left\{O_{j_i}(s_{j_i}), \mb x_{j_i}(s_{j_i}), \mb y_{j_i}(s_{j_i}), \mb z_{j_i}(s_{j_i})\right\}$, so that the $\mb{z}_{j_i}(s_{j_i})$ axis is tangent to the curve and define $F_{j_i} \equiv F_{j_i}(l)$; $F_{1_0}, \ F_{2_0}, \dots F_{N_0}$ are the ground frames of each of the $N$ tools, assumed to have a coincident origin $O \equiv O_{j_1}(0) \ \forall j$. We define a world reference frame $W = \left\{O, \mb x_0, \mb y_0, \mb z_0\right\}$ so that $p_{j_0} = 0$ is the distance between the frame $F_{j_0}$ and $W$ and $R_{j_0} = \text{rot}_{\mb z_{j_0}}(\theta_j)$ their relative orientation. The axial rotation of the $j$-th tool (see Fig. \ref{fig:model}), $\text{rot}_{\mb z_{j_0}}(\theta_j)$, of the angle $\theta_j$ around the axis $\mb z_{j_0}$ can be controlled as described in Section \ref{sec:planning}. 

The bending of each segment is defined by the local bending vector $\pmb{\gamma}_{j_i} \in \mathbb{R}^3$, with respect to the local reference frame $F_{j_i} \equiv F_{j_i}(l) = \left\{O_{j_i}(l), \mb x_{j_i}(l), \mb y_{j_i}(l), \mb z_{j_i}(l)\right\}$; $\pmb{\gamma}_{j_i}$ is described as the deflection along the 3 principal directions. Here $j = 1, 2, \dots, N$ with $N$ number of tools and $M_j$ is the number of segments for each tools, $i = 1, 2, \dots, M_j$.

In this work, the elongation of one tool with respect to the others is described as increase in the number of segments in each tool. $M_{j - 1} - M_{j}$ describes the amount of extension of the tool $j - 1$ (innermost) with respect to $j$ (outermost).

The position and orientation of each frame $F_{j_i}(s_{j_i})$ along the centerline of the $j$-th tool can be written with respect to the world frame $W$ \cite{Pittiglio2023ClosedRobots}, as
\begin{subequations}
\label{eq:kinematics}
\begin{equation}
\label{eq:kinematics:position}
\mb{p}_{j_i}(s_{j_i})  = \mb{p}_{j_{i-1}}(l) + \mb{R}_{j_{i-1}}(l) \: \text{exp} \left(\left[\pmb{\gamma}_{j_i}\frac{s_{j_i}}{l}\right]^\wedge \right)  \boldsymbol{e}_3
\end{equation}
\begin{equation}
\mb{R}_{j_i}(s_{j_i}) = \mb{R}_{j_{i-1}}(l) \: \text{exp} \left(\left[\pmb{\gamma}_{j_i}\frac{s_{j_i}}{l}\right]^\wedge \right)
\end{equation}
\end{subequations}
for the respective position and rotation.

To describe the static equilibrium of each tool we write the Lagrangian
\begin{equation}
    L_j = T_j - U_j = -U_j,
\end{equation}
where we assumed the kinetic term $T_j$ is negligible, i.e. $T_j = 0$. The potential energy can be written as 
\begin{eqnarray}
\label{eq:potential}
U_j & = & \sum_{i = 1}^{M_j} \frac{1}{2} \int_{0}^{l} -\pmb \gamma_{j_i}\T \mb k \pmb \gamma_{j_i} - 2 m_i \mb g\T \mb p_{j_i}(s_{j_i}) \de s_{j_i} \nonumber \\
& = & -\frac{1}{2} \bs \gamma_j\T \text{diag}(\underbrace{\mb k l, \ \mb k l, \ \dots, \ \mb k l}_{M_j \ \text{times}}) \bs \gamma_j - \nonumber \\
&& \sum_{i = 1}^{M_j} m_i \mb g\T \int_{0}^{l}\mb p_{j_i}(s_{j_i}) \de s_{j_i} \nonumber ,
\end{eqnarray}
 with $m_i$ mass density of the $i$-th tool and $\mb g$ gravitational acceleration.

We consider the mechanical stiffness of the tool
\begin{equation}
\mb k = \text{diag} \left(1, \ 1, \ \frac{1}{2(\nu + 1)} \right)\frac{E A}{l}
\end{equation}
with $E$ \emph{Young's modulus}, $A$ second moment of area and $\nu$ \emph{Poisson's ratio}. The first two elements of $\mb k$ describe the bending stiffness, while the last one accounts for the torsional stiffness, since $\mb{z}$-axis is defined to be tangent to the backbone curve.

Since our tools are completely passive, we have no external forces, apart from the ones arising from their interactions and their contact with the anatomy. These are going to be treated as constraints, as discussed in the following sections. The equilibrium of each tool, treated as independent flexible bodies, can be found by solving the minimization problem
\begin{equation}
\label{eq:single_solution}
    \min_{\pmb \gamma} U_j, \ i = 1, 2, \dots, M_j
\end{equation}
with $\pmb{\gamma} = \left(\pmb \gamma_{1_1} \ \cdots \pmb \gamma_{1_{M_1}} \cdots \pmb \gamma_{N_1} \cdots \pmb \gamma_{N_{M_N}} \right)\T$

\subsection{Telescoping Tools Constraints}
\label{sec:model:tools}
From the statics of each independent tool, described above, we can find their shape depending on their mechanical stiffness and under gravitational load. To consider their interaction, we constrain their centerline to be coincident for the overlapping segments. For the $j$-th tool, we can find number of segments overlapping with the $k$-th tool as $L_{j, k} = \|M_j - M_k\|$. Since we modeled the tools having a common ground frame $W$, tools $j$ and $k$ overlap for the first $L_{j, k}$ segment. The segments overlap if they satisfy the constraints
\begin{equation}
\label{eq:tele_constraints}
    \mb C^{(i)}_{j, k} = \mb p_{j_i}(l) - \mb p_{k_i}(l) = \mb{0} \in \mathbb{R}^3, \ i = 0, 1, \dots, L_{j, k}
\end{equation}
We can build the overall constraint vector $\mb C$, by stacking together all the constraints between tools and removing repetitions arising from multiple overlapping elements.

\subsection{Anatomical Constraints}
\label{sec:model:anatomy}
From preoperative imaging, we can segment the anatomy and convert its walls into a closed triangulated surface, as depicted in Fig. \ref{fig:model}. The $k$-th triangle has a center point $\mb s_k \in \mathbb{R}^3$ and a normal $\mb h_k \in \mathbb{R}^3$ associated with it. The normals to each of the $D$ triangles are chosen to point inwards, with respect to the anatomy. To find whether a point along the centerline of any of the tools $\mb p_{j_i}(l)$ is inside the anatomy, we find the nearest triangle 
\begin{equation}
    k_{j,i} = \argmin_{k} \|\mb p_{j_i}(l) - \mb s_k\|, \ k = 1, 2, \dots, D.
\end{equation}

and compute the projection of vector between the triangle and point along the centerline
\begin{equation}
    d_{j_i} = (\mb p_{j_i}(l) - \mb s_{k_{j,i}}) \cdot \mb h_{k_{j,i}}.
\end{equation}
When the projection is positive the point is inside the anatomy, by definition. We can stack together the the projected distances $d_{j_i}$ in a vector of constraints 
\begin{equation}
\label{eq:anat_constraints}
    \mb S = \left(d_{1_1} \ \cdots d_{1_{M_1}} \ \dots d_{N_{M_N}}  \right) \T .
\end{equation}

\subsection{Static Equilibrium Solution}
\label{sec:model:solution}
To find the overall configuration of the telescoping tools, we combine (\ref{eq:single_solution}), (\ref{eq:tele_constraints}) and (\ref{eq:anat_constraints}). We write the constrained static equilibrium as
\begin{eqnarray}
\label{eq:solution}
        & \displaystyle \min_{\pmb \gamma} & \mb U = \left(U_1 \ U_2 \cdots U_N \right)\T \\
         & \text{subject to } & \mb C = 0 \nonumber \\
        & & \mb S \geq 0 \nonumber
\end{eqnarray}

\section{RRT Planner for Neuroendovascular Tools}
\label{sec:planning}
From pre-operative imaging, obtained from common imaging modalities such as \gls{cta} or \gls{mra}, and knowledge of the properties of the catheterization tools, we plan the contact-aware motion inside the anatomy. In this work, we assume a 3D segmentation of the pre-operative images is provided, commonly from the aorta up (Fig. \ref{fig:platform}) \cite{vanderZijden2019CurrentClinician}.

The control variables for our neuroendovascular tools are the number of segments we insert for the $j$-th tool, $M_j$, and its axial rotation $\theta_j$, defined in the previous section.

For a certain configuration $V(t) = \left\{M_j(t), \theta_j(t), j = 1, 2, \dots N\right\}$, we can solve the optimization problem in (\ref{eq:solution}) and obtain a configuration for the tools $\pmb \gamma(t)$ which guarantees: (i) the tools act telescopically; (ii) the tools are within the anatomy or in contact with its boundaries. By solving the direct kinematics in (\ref{eq:kinematics}), we obtain the position of the tip of the innermost tool $\mb p_{1_{M_1}}(l, t) \triangleq \mb p(t)$. 

From an initial known configuration $V(0)$, we aim to find the series of configurations $V(1), V(2), \dots, V(L), \ L \geq 0$ such that $\| \mb p(L) - \mb q\| < \epsilon$, with $\mb q \in \mathbb{R}^3$ target and a small enough $\epsilon$. The target is defined by the surgeon from a 3D pre-operative map of the anatomy.

To find the optimal series of configurations, we developed a \emph{contact-aware} \gls{rrt} algorithm to build the graph $G$, as described in Algorithm \ref{alg:rrt}.

\begin{algorithm}
\caption{Contact-aware RRT}\label{alg:rrt}
$t \gets 0$ \\
$V(t) \gets$ initial configuration \\
$G \gets$ INIT($V(t)$) \\
$\mb p(t) \gets$ DIR\_KIN($V(t)$)\\
\While{$1 < t \leq L$}{
  \If{$\| \mb p(t) - \mb q\| > \epsilon$}{
    $V_{rand} \gets$ RAND\_CONF() \\
    $V_{near} \gets$ NEAREST\_VERTEX($V_{rand}, G$) \\
    $V_{new} \gets$ STEER($V_{near}$, $V_{rand}, \Delta V$) \\
    $G.$ ADD\_VERTEX($V_{new}$) \\
    $G.$ ADD\_EDGE($V_{new}$, $V_{near}$) \\
    $\mb p(t) \gets $ DIR\_KIN($V(t)$)\\
  }
}
\KwResult{$G$}
\end{algorithm}

Once the graph $G$ is built, we search for the shortest path from $V(0)$ to $V(L)$.

The function INIT($V(t)$) adds the vertex $V(t)$ to the graph and initializes the graph with no edges. DIR\_KIN($V(t)$) finds the solution to the problem (\ref{eq:solution}) and computes the direct kinematics to find the tip of the innermost tool $\mb p(t)$, via $(\ref{eq:kinematics})$. We generate a random configuration within the maximum elongation of each tools with the function RAND\_CONF() and find the nearest vertex in the graph as
\begin{equation}
    \text{NEAREST\_VERTEX}(V_{rand}, G) := \argmin_{V \in G} \|V - V_{rand} \|.
\end{equation}

Given the limits in the control effort $\Delta V = \left(\Delta M_1 \ \Delta \theta_1 \ \Delta M_2 \ \Delta \theta_2 \ \cdots \ \Delta M_N \ \Delta \theta_N \right) =\left(\Delta V_1 \ \Delta V_2 \ \dots \Delta V_{N} \right)  \in \mathbb{R}^{2 N}$ for each of the control variables, we steer between nodes as
\begin{eqnarray}
    && \text{STEER}(V_{near}, V_{rand}, \Delta V) := \\
    && V_{{near}_i} + \text{sign}(V_{{rand}_i} - V_{{near}_i})\Delta V_{i}, \ i = 1, 2, \dots, N \nonumber
\end{eqnarray}

The functions $G.$ADD\_VERTEX($V_{new}$) and $G.$ADD\_EDGE($V_{new}$, $V_{near}$) add $V_{new}$ vertex to the graph $G$ and edge connecting $V_{near}$ to $V_{new}$, respectively.

\section{Experimental Validation}
\label{sec:experiments}
The first task in any neuroendovascular surgery is reaching the brain vessels from the aorta through the carotid arteries. In clinical practice, multiple trials may be necessary for a surgeon to identify an optimal path that bypasses certain branches and reaches the target branch. 

\begin{figure}[t] 
    \centering
    \includegraphics[width=\columnwidth]{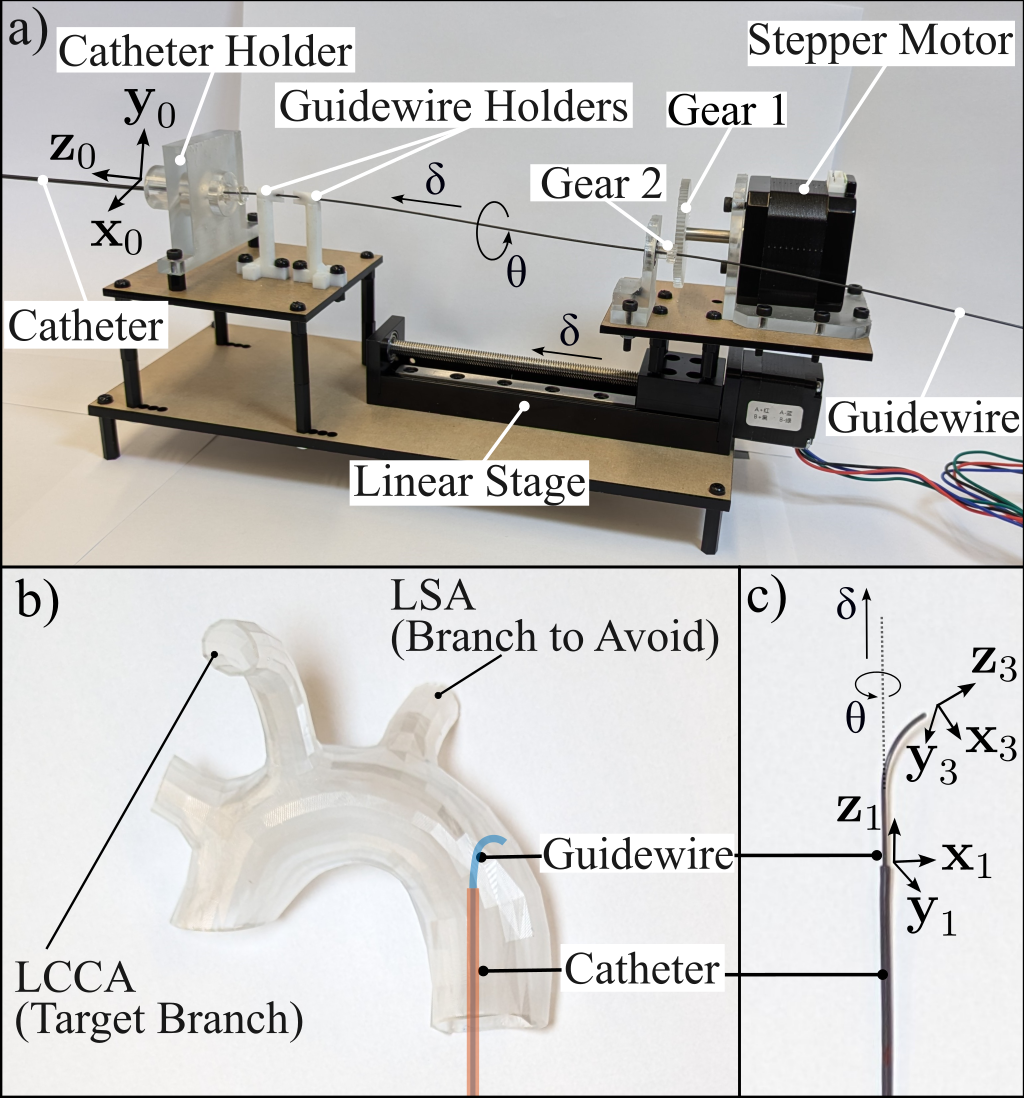}
    \caption{Components of the experimental setup. a) Actuation system for controlling guidewire translation and rotation; b) Guidewire and Catheter system inside anatomy; c) Tools configuration parameters.}
    \label{fig:actuation_system}
\end{figure}

For validation of the planner presented in Section \ref{sec:planning}, an experimental setup was designed to simulate this navigation task. We built a robotic platform (Fig. \ref{fig:actuation_system}a), described in Section \ref{subsect:actuation_system}, to actuate the tools as commanded by the planner. The objective is to guide the tools through a model of the aorta and into the \gls{lcca}, avoiding the \gls{lsa}. A phantom model of the aorta (Fig. \ref{fig:actuation_system}b), from blenderkit library (blenderkit.com), was printed using a stereolithography (SLA) printer. The tools used in this setup (Fig. \ref{fig:actuation_system}c) include a Radifocus Glidewire Guidewire (Tool 1) with a 0.035\,inch diameter, 260\,cm length, and a 3\,cm angled tip, and a straight Terumo Radifocus Glidecath Catheter (Tool 2) with a 1.70\,mm diameter and 100\,cm length. The guidewire used for the experiments consists of a nitinol core, and for modeling purposes, the Young's modulus was taken primarily as that of NiTi (70\,GPa), as it dominates over the guidewire’s coating material. The catheter's Young's modulus was assumed negligible, given its much more flexible nature when compared to the NiTi guidewire. 

\begin{figure}[t] 
    \centering
    \includegraphics[width=\columnwidth]{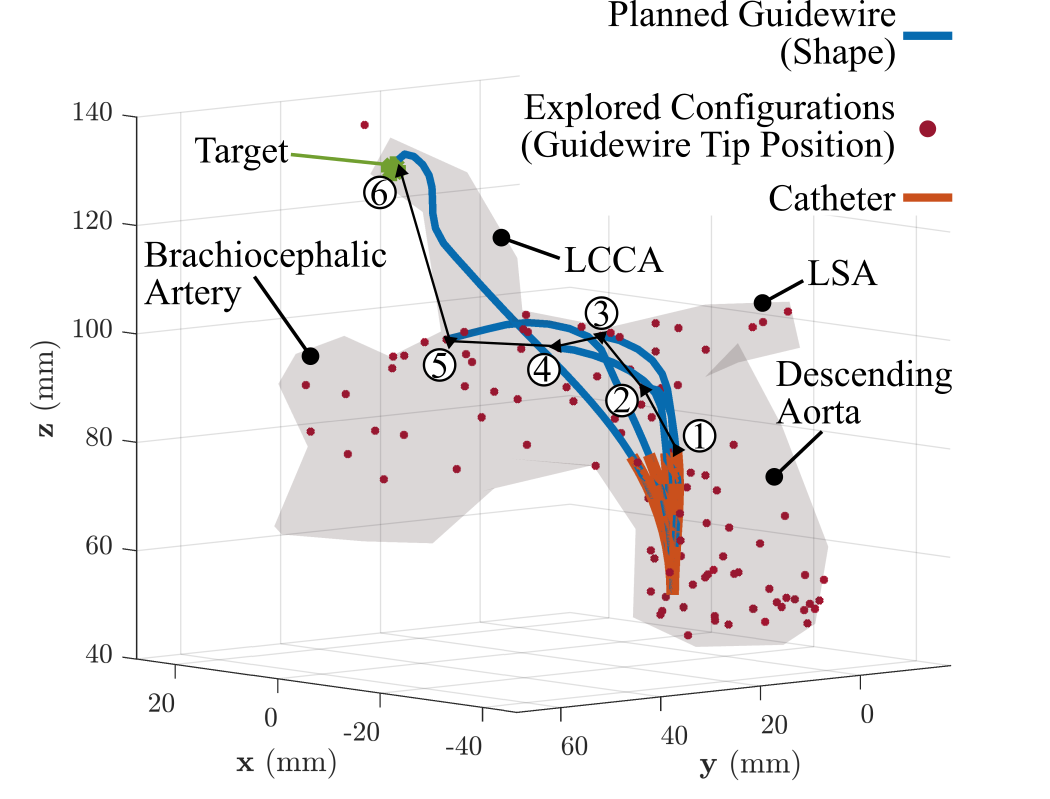}
    \caption{Contact-aware planned path from base of descending aorta to \gls{lcca}, avoiding \gls{lsa}. \circled{1} Start configuration, \circled{6} End configuration.}
    \label{fig:plan}
\end{figure}

The planner was applied to this setup to identify a feasible path for the guidewire from the initial configuration to the target location (the \gls{lcca}) while avoiding the \gls{lsa}. 
The results are shown in Fig. \ref{fig:plan}. We ran the algorithm for 100 configurations (Explored Configurations in Fig. \ref{fig:plan}) and found the shortest path from the initial configuration \circled{1}, where both tools are inserted of 12.5\,mm inside the aorta, to the target \circled{6} in the \gls{lcca}. For this task, the planner only required actuation of the guidewire. Hence, we built a robot for only guidewire actuation, as described in Section \ref{subsect:actuation_system}.

\subsection{Actuation System}

\label{subsect:actuation_system}
Fig. \ref{fig:actuation_system}a, illustrates an actuation system designed to control the translation and rotation of the guidewire. For translation of the guidewire along its axis, a 100\,mm linear stage with a T6$\times$1 lead screw driven by Nema 11 stepper motor is used. For the rotation of the guidewire around its axis, a gear assembly was designed, including a driver gear and a driven gear to transmit controlled rotational motion to the guidewire. The driver gear (Gear 1), connected to a Nema 17 stepper motor, has a pitch diameter of 41\,mm with 80 teeth, while the driven gear (Gear 2) has a pitch diameter of 7\,mm and 12 teeth, allowing the rotational control. 

The catheter is held in place using a catheter holder, while the two guidewire holders keep the guidewire stable and aligned, preventing it from moving off-axis. These holders help keep the guidewire straight and avoid any twisting, coiling, or folding back of the guidewire into the catheter.

An arduino Uno REV3 along with a CNC shield V3 extension board and A4988 stepper motor drivers are used are used to control the actuation system.

As discussed in Section \ref{sec:planning}, the planner commands number of segments inserted for the guidewire, $M \triangleq M_1$. Given the length of each infinitesimal element $l$, we compute the insertion of the linear stage $\delta = M \cdot l$.

In this experiment, $l$ was set to $1.25$\,mm. Thus, each insertion step defined by the planner corresponds to an insertion length of $\delta = 1.25$\,mm in the physical setup.

In addition to insertion, the planner specifies rotational adjustments of the guidewire along its longitudinal axis $\theta \triangleq \theta_1$, as shown in Fig. \ref{fig:actuation_system}c. This parameter ensures that the guidewire is oriented correctly to follow the planned path through the aorta phantom. For the experiments, it was assumed that the rotation produced by the gear mechanism is fully transferred to the guidewire base.

\subsection{Results}
\begin{figure}[t] 
    \includegraphics[width=\columnwidth]{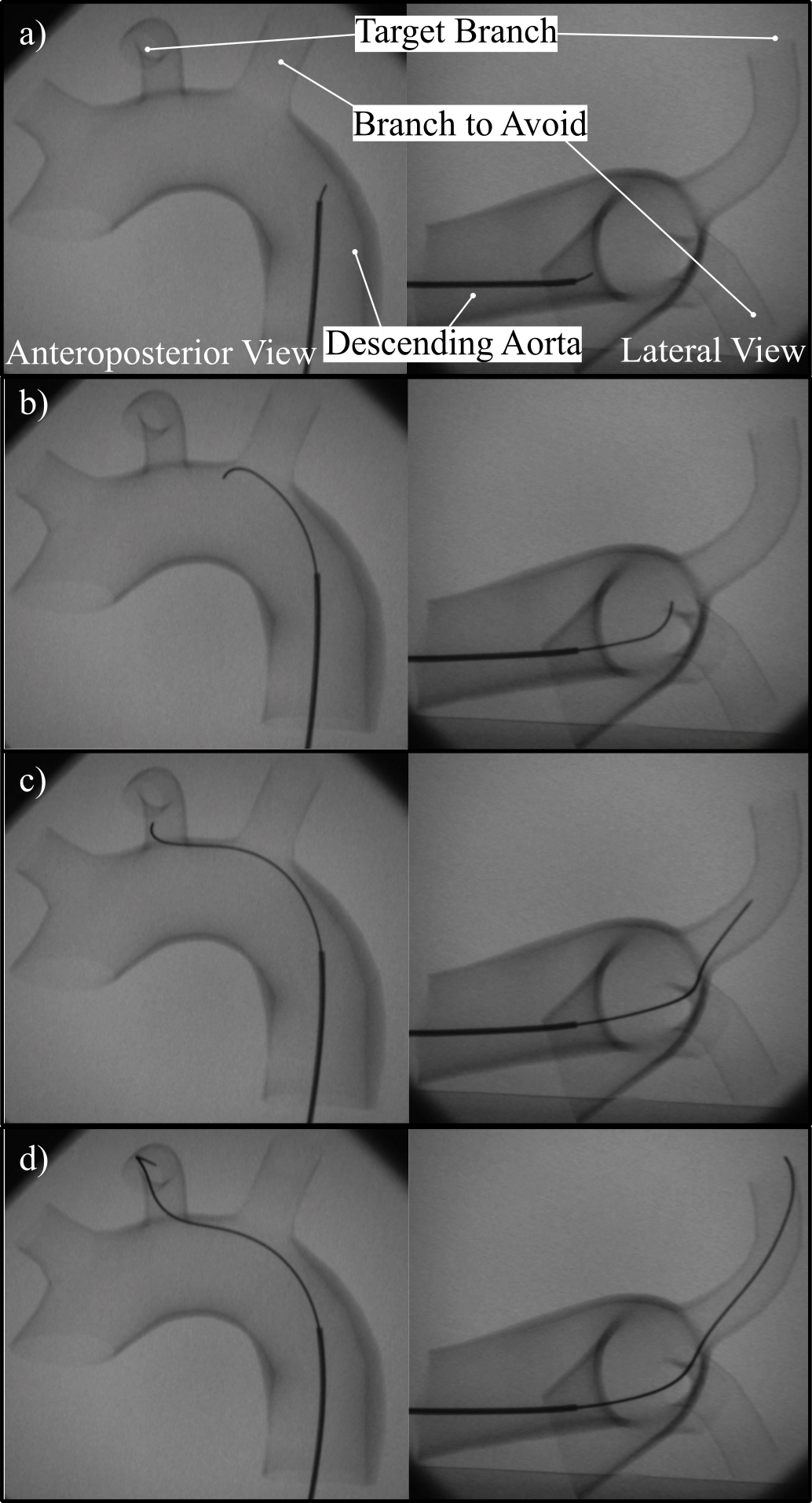}
    \caption{Guidewire navigation in the phantom of the arteries in antero/posterior and lateral views; a) initial configuration; b) guidewire avoiding \gls{lsa}; c) guidewire entering \gls{lcca}; d) end configuration.}
    \label{fig:results}
\end{figure}
The pre-operative path generated by the planner (see Fig. \ref{fig:plan}) was executed by the actuation unit in Fig. \ref{fig:actuation_system}a in the anatomy in Fig. \ref{fig:actuation_system}b. Validation was performed by observing the guidewire's navigation through the aorta, past the \gls{lsa} (branch to avoid), to reach the \gls{lcca} (target branch). Navigation was considered successful when the guidewire reached the end of the section of the \gls{lcca} represented in the phantom.

We ran the experiment 50 times with 100\% success rate. We show two of the trials in the Supplementary Video. We captured the navigation using single plane fluoroscopy and repeated it twice to collect antero-posterior (A/P) and lateral view images. Screenshots from the videos are reported in Fig. \ref{fig:results}.

In both view sequences, the guidewire begins from the descending aorta in its initial configuration (Fig. \ref{fig:results}a). The guidewire proceeds along the descending aorta and avoids to enter the \gls{lsa}, as seen in Fig. \ref{fig:results}b. By guiding itself along the walls of the anatomy, the guidewire reaches the based of the \gls{lcca}, shown in Fig. \ref{fig:results}c, via an axial rotation, it enters the \gls{lcca} and extends into the artery (Fig. \ref{fig:results}d) until its end, i.e. the target area. In all 50 independent executions, the guidewire follows the planned path using contact-aware motion planner presented in Section \ref{sec:planning}, from the aorta to the \gls{lcca} while successfully avoiding the \gls{lsa}.

\begin{figure}[t] 
    \centering
    \includegraphics[width=0.65\columnwidth]{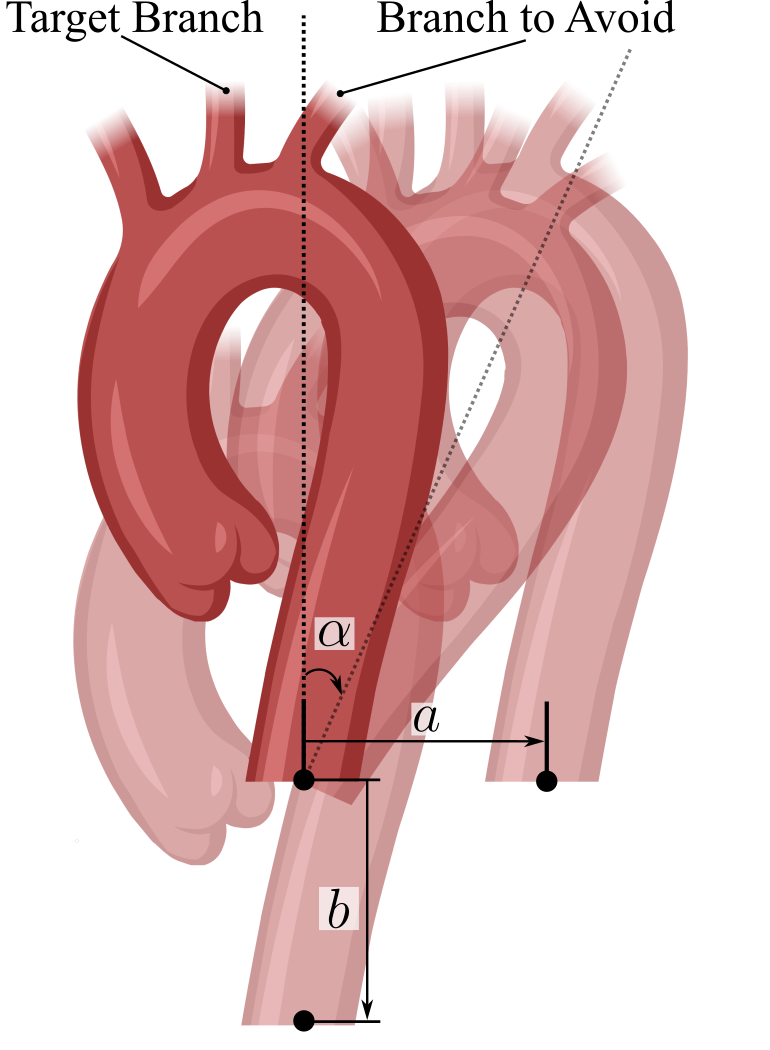}
    \caption{Configurations of the aorta considered to test the robustness of the pre-operative plan.}
    \label{fig:robustness}
\end{figure}

\begin{figure}[t] 
    \centering
    \includegraphics[width=\columnwidth]{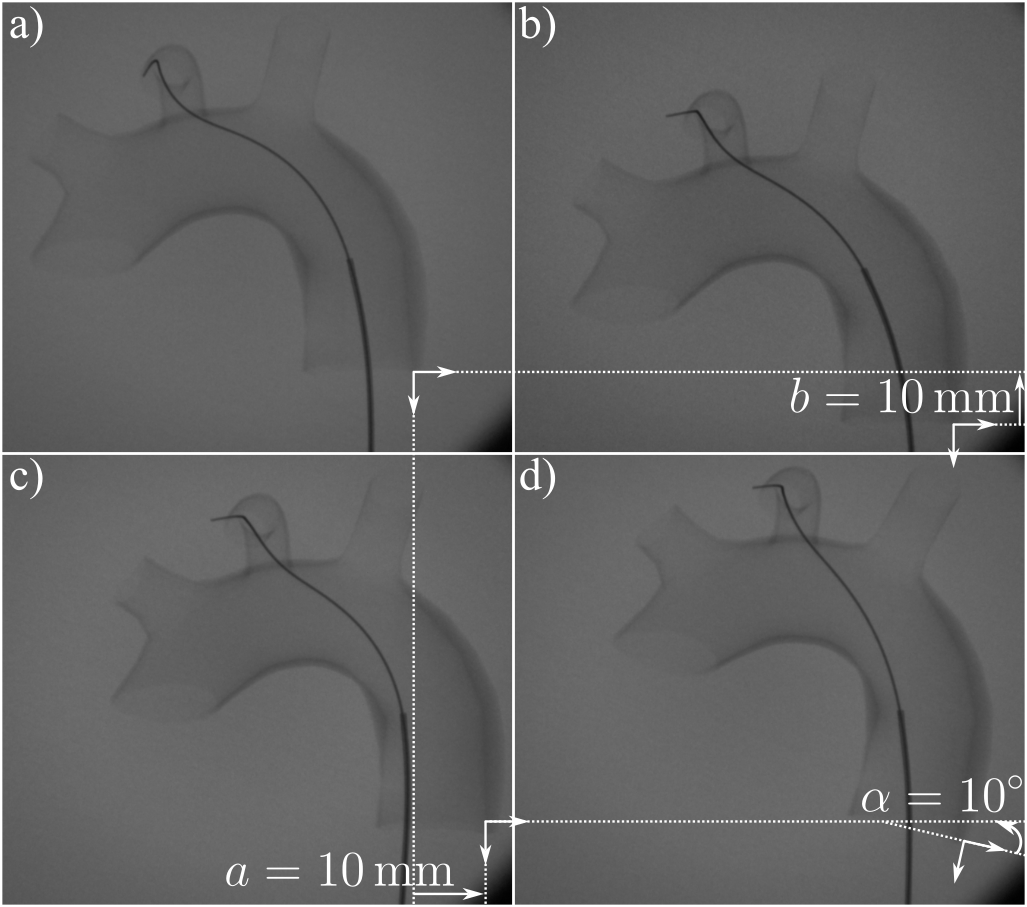}
    \caption{Results of planner robustness towards anatomical motion.}
    \label{fig:robustness_flou}
\end{figure}

In the clinical settings, minor misalignment of the guidewire or slight movements of the anatomy can occur, and might affect the planner's performance. To test the robustness of the planner under these conditions, we ran the experiments while varying the placement of the phantom with respect to it's nominal configuration (Fig. \ref{fig:robustness_flou}a) - i.e. the one the planner was run on. This included adjustments in axial displacement, lateral displacement, and rotation of the phantom on the coronary plane. The tools where fixed in placed to emulate misplacement with respect of the anatomy or anatomical motion.

The configurations we considered are shown in Fig. \ref{fig:robustness}. They include displacement and rotation on the coronal plane, defined as: $a\in \mathbb{R}$ for lateral displacement, $b\in \mathbb{R}$ for axial displacement, and  $\alpha\in \mathbb{R}$ as the rotation orthogonal to the coronal plane. These parameters represent common misalignments that can occur due to slight guidewire shifts or patient movements. The planner demonstrated robustness against these parameters with tolerance ranges of $a \in [-5, +10]$\,mm for lateral displacement, $b \in [-7.5, +10]$\,mm for axial displacement, and $\alpha \in [-5, +10]^\circ$. 

In Fig. \ref{fig:robustness_flou}, we report A/P fluoroscopy scans to show the largest positional and angular displacements which does not affect the planner's performances: $b = 10$\,mm (Fig. \ref{fig:robustness_flou}b); $a = 10$\,mm (Fig. \ref{fig:robustness_flou}c); $\alpha = 10^\circ$ (Fig. \ref{fig:robustness_flou}d). Despite these variations in position and orientation, the guidewire consistently reached the target branch (\gls{lcca}) while avoiding the undesired branch (\gls{lsa}). 
These findings indicate that the path provided by the planner can effectively adapt to misalignment of the catheter and movements of the anatomy.

\section{Conclusions}
\label{sec:conclusions}
In the present work, we introduced a model-based contact-aware planner for autonomous catheterization interventions for acute ischemic stroke treatment. The method was developed for telescoping pre-bent tools currently used for neuroendovascular interventions, which are unable of tip steering and advance by interacting with the walls of the vessels. We present an actuation system able of actuating insertion and rotation of such telescoping tools and experimental validation of successful navigation from the base of the descending aorta to the \gls{lcca}.

For this first task, necessary to further access other \glspl{lvo}, we only required planning and actuation of the innermost tool: the guidewire. We demonstrate that the planner can successfully avoid the \gls{lsa}, which it would naturally enter without appropriate path planning, and can navigate to the \gls{lcca} 100\% of times over 50 trials. The planner was also proven robust towards rotations of the aorta of up to $10^\circ$, and displacement of 10\,mm on the coronal plane, which may arise in the clinical settings.

Future work will focus on further navigation to the large vessels within the cerebral cavity, by actuating pre-bent catheters in combination with guidewires. Real-time fluoroscopic images will be used for model-based closed-loop control, using the model presented in this paper.


\bibliographystyle{IEEEtran}
\bibliography{references}											
\end{document}